\documentclass[conference]{IEEEtran}
\IEEEoverridecommandlockouts
\usepackage{cite}
\usepackage{amsmath,amssymb,amsfonts}
\usepackage{hyperref}
\usepackage{algorithmic}
\usepackage{graphicx}
\usepackage{textcomp}
\usepackage{xcolor}
\usepackage{comment}
\usepackage{multicol, blindtext}
\usepackage{mwe}
\usepackage{float}
\usepackage{multirow}
\usepackage{makecell}
\usepackage{subfig}
\usepackage{verbatim}
\usepackage[export]{adjustbox}
\usepackage{enumerate}
\usepackage{fixltx2e}
\usepackage{longtable,rotating}
\usepackage{threeparttable}
\usepackage{blindtext}
\usepackage[linesnumbered,ruled,vlined]{algorithm2e}
\SetKwInput{KwInput}{Input}         
\SetKwInput{KwOutput}{Output}              
\SetKwRepeat{Do}{do}{while}%

\def\BibTeX{{\rm B\kern-.05em{\sc i\kern-.025em b}\kern-.08em
    T\kern-.1667em\lower.7ex\hbox{E}\kern-.125emX}}
\begin{document}

\title{EnsembleNTLDetect: An Intelligent Framework for Electricity Theft Detection in Smart Grid\\}
\author{{Yogesh Kulkarni\IEEEauthorrefmark{1}, Sayf Hussain Z\IEEEauthorrefmark{2}, Krithi Ramamritham\IEEEauthorrefmark{3}, Nivethitha Somu\IEEEauthorrefmark{4}}\\
    {\IEEEauthorrefmark{1}Department of Computer Engineering, Pune Institute of Computer Technology, Pune, India}\\
    {\IEEEauthorrefmark{2}Department of Computer Science \& Engineering, College of Engineering, Guindy, Anna University, Chennai, India}\\
    {\IEEEauthorrefmark{3}Robert Bosch Centre for Data Science \& Artificial Intelligence, IIT Madras, Chennai, India}\\
    {\IEEEauthorrefmark{4}Department of Electrical Engineering, IIT Bombay, Mumbai, India}\\
    \{yogeshpict, sayfhussain.10, ramamrithamk, nivethithasomu\}@gmail.com
    }

\maketitle
\let\thefootnote\relax\footnotetext{Accepted at the 2nd Workshop on Large-scale Industrial Time Series Analysis at the 21st IEEE International Conference on Data Mining (ICDM), 2021.}

\begin{abstract}
Artificial intelligence-based techniques applied to the electricity consumption data generated from the smart grid prove to be an effective solution in reducing Non Technical Loses (NTLs), thereby ensures safety, reliability, and security of the smart energy systems. However, imbalanced data, consecutive missing values, large training times, and complex architectures hinder the real time application of electricity theft detection models. In this paper, we present EnsembleNTLDetect, a robust and scalable electricity theft detection framework that employs a set of efficient data pre-processing techniques and machine learning models to accurately detect electricity theft by analysing consumers' electricity consumption patterns. This framework utilises an enhanced Dynamic Time Warping Based Imputation (eDTWBI) algorithm to impute missing values in the time series data and leverages the Near-miss undersampling technique to generate balanced data.\\
Further, stacked autoencoder is introduced for dimensionality reduction and to improve training efficiency. A Conditional Generative Adversarial Network (CTGAN) is used to augment the dataset to ensure robust training and a soft voting ensemble classifier is designed to detect the consumers with aberrant consumption patterns. Furthermore, experiments were conducted on the real-time electricity consumption data provided by the State Grid Corporation of China (SGCC) to validate the reliability and efficiency of EnsembleNTLDetect over the state-of-the-art electricity theft detection models in terms of various quality metrics.
\end{abstract}


\section{Introduction}\label{Intro}
With the apparent increase in the global electricity demand, setting up new generation plants is often a difficult and tedious process due to several constraints enforced by the pollution control and environmental conservation policies \cite{b1}. The electricity loss during the generation, transmission, and distribution of electricity in the power grid is a critical challenge faced by the power utilities across the globe. Such electricity losses can be classified as \cite{b2, b3}: (1) \textbf{Technical Losses (TLs):} occurs during transmission. e.g., dissipation of power in resistors, transmission lines, transformers, etc. and (2) \textbf{Non-Technical Losses (NTLs):} the clear difference between the total loss and the TLs. e.g., meter tampering, electricity theft, faulty meters, billing errors, and other irregularities to evade payment to the utility company by the consumers.
Among these, NTLs affect the utilities’ revenue and the nation’s economy with their drastic impact on the quality of power supply, increased load on the power stations, and high tariffs on genuine consumers. Developed countries like U.S and U.K experience NTLs but are not as large as developing countries in Asia, and Africa \cite{b4}. In particular, electricity theft, defined as the illegal use of electricity with an intention to avoid billing charges, forms a major part of the NTLs \cite{b5}. Electricity theft is a complex research problem with several influential parameters like socio-economic, regional, infrastructure, corruption, managerial, etc. \cite{b6}. In general, electricity theft occurs at (i) \textbf{Consumers:} energy tapping and meter tampering, (ii) \textbf{Utility:} billing inaccuracies, and (iii) \textbf{Grid:} bypass meters. The electricity theft at the grid and consumer level results in serious implications for the utilities since it affects their profit and economic wellness of the nation through reduced investments in the power sector, high financial loss (around \$4.5-25 billion per YEAR), electrocution deaths, and frequent power outages with overloaded generation units \cite{b6}. Moreover, it is tough for the utilities to detect and confirm electricity theft in domestic, commercial and industrial establishments, rural areas and large cities through on-site inspections, an inefficient and expensive manual process. 

The advent of Advanced Metering Infrastructures (AMIs) in smart grids accompanied with low-cost smart meters enables two-way communication between the customer and the utility provider for the accounted metering and billing process through fine-grained electricity consumption data \& periodic information flow on energy supply and demand. Such advancements accompanied by the massive electricity consumption data have instigated the researchers and the utilities to apply IoT, Big Data, and Artificial Intelligence techniques for the design of efficient and intelligent electricity theft detection mechanisms and accurate utility operations \cite{b7}. The design of a reliable and efficient electricity theft detection mechanism aids the utilities to enforce legal actions on illegal consumers, achieve expected profit and future investments in the power sector for reliable \& secure power services. In this way, several machine learning and deep learning techniques have been profoundly applied to energy research problems such as energy trading, virtual power plant, energy consumption monitoring and control, and electricity theft for the design of future intelligent energy networks \cite{b8}. The state-of-the-art electricity theft detection approaches can be widely classified into three, namely (i) \textbf{State based detection approaches:} monitors the state of the grid and smart meters through RFIDs and sensors; high cost of deployment and maintenance, (ii) \textbf{Game theory-based detection approaches:} provides a low-cost solution through modelling a game between the consumers and utility provider; determining utility functions of the participants (consumers, distributors, regulators, etc.) is complex, and (iii) \textbf{Artificial intelligence-based detection approaches:} clustering and classification approaches proves to be an cost-effective and reliable solution with the inherent ability of massive electricity consumption data provided by
the tamper-proof smart meters to understand the consumer electricity consumption profiles. Owing to the massive electricity consumption data and advanced artificial intelligence approaches, the literature analysis of this paper is confined to the artificial intelligence based electricity theft detection models.

\begin{figure*}[htbp]
\centering
  \includegraphics[width=160mm]{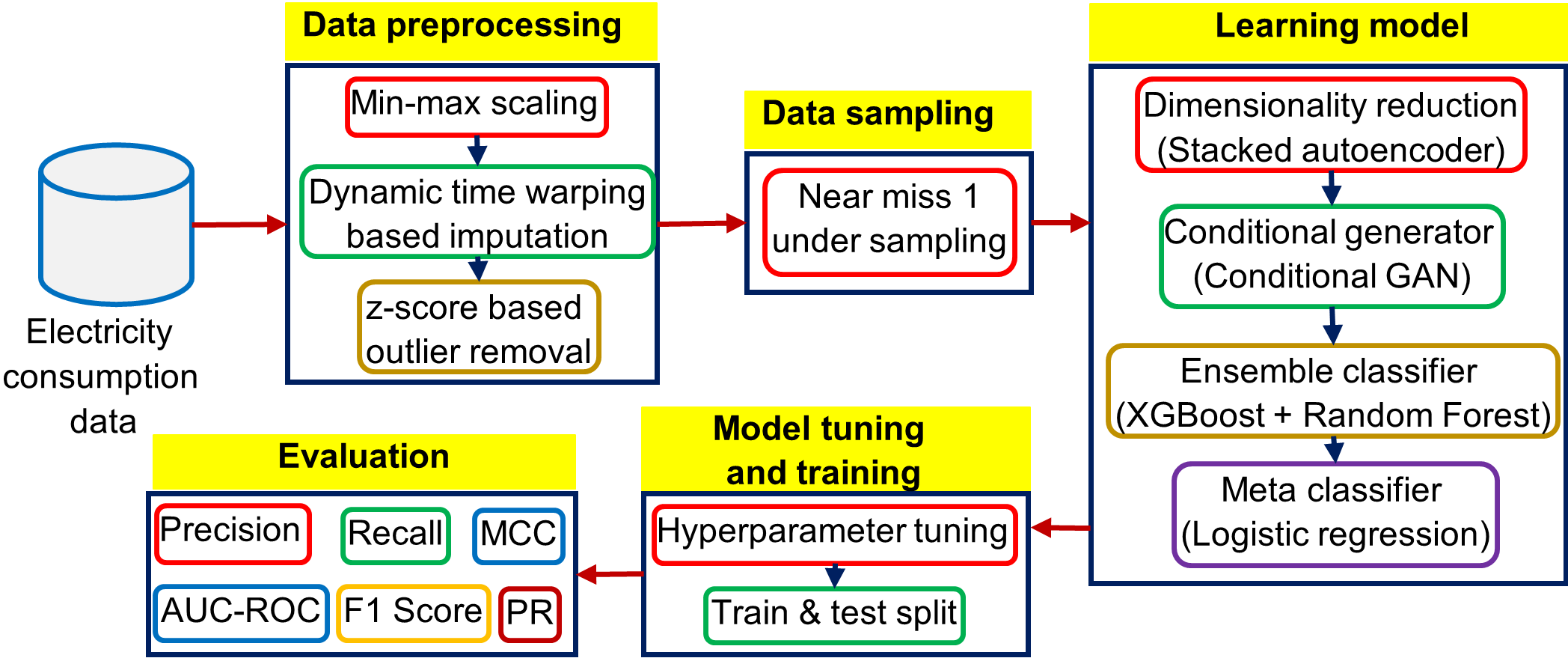}
  \caption{Architecture of EnsembleNTLDetect}
  \label{complete_block}
\end{figure*}

Support Vector Machine (SVM) is the most commonly used technique for electricity theft detection to achieve a high detection rate and fewer false alarms. Certain aspects of the electricity consumption data such as historical consumption data (location, seasonality, and category), load profile information, identification of consumers with a high probability of abnormal behaviour, and high dimensional data have been explored well using SVMs \cite{b9}, Genetic algorithm-based SVM \cite{b5}, fuzzy-based SVMs \cite{b10}, and PCA based SVMs \cite{b11}. Electricity thieves have also been identified by analyzing their load profiles at different hierarchies of the power grid (transmission, distribution, and consumer) using hybrid SVM models such as decision tree-based SVMs \cite{b12}, decision trees-k-nearest neighbour SVMs \cite{b13}, Extreme learning machine (ELM), online sequential ELMs \cite{b14}, and even multi-class SVMs \cite{b15}. Studies in \cite{b16,b17} have carried out a detailed comparative analysis of machine learning models to detect NTLs. Regression and distance-based models like AutoRegressive Moving Average (ARMA) \cite{b18}, Nonlinear AutoRegressive with eXogenous input (NARX) \cite{b19}, linear regression \cite{b20}, \textit{k}-means (KM) clustering-based ANNs \cite{b16}, fuzzy C-means clustering \cite{b10}, Extreme Gradient Boosting \cite{b21} and Optimum Path Forest (OPF) \cite{b22} were employed to detect NTLs with the detection accuracy between 77\%-97\%. The inherent ability of deep learning architectures to handle real-time high dimensional smart meter data and automated feature extraction capabilities have led to the development of various single and hybrid deep learning-based electricity theft detection models using Convolutional Neural Networks (CNN) \cite{b23, b24}, Long Short Term Memory (LSTM) \cite{b25}, Self-organizing Map (SOM) and Multilayer Perceptron Artificial Neural Network (MP-ANN) \cite{b26}, and Particle Swarm Optimization based Stacked Sparse Denoising Auto Encoder (SSDAE) \cite{b27}. Notable contributions on electricity theft detection have used Kullback–Leibler divergence \cite{b28}, a combination of state estimation, multivariate control charts and A* path search algorithm \cite{b29}, applied self-organizing maps \cite{b30}, and 
undersampling boosting algorithms \cite{b31}.

The challenges in the state-of-the-art electricity theft detection models such as imbalanced nature of the data, consecutive missing values in the time series data, capturing the seasonal trends while imputing missing values, complex architectures, high training time needs to be taken care of for the design of an efficient and robust electricity theft detection model. This paper formulates the identified challenges as following research questions: (i) How to handle large gaps, i.e., consecutive missing values, in time series data with high seasonal trends effectively? (ii) What is the impact of undersampling techniques on generating a balanced electricity consumption dataset without information loss? (iii) How to handle the high dimensional electricity consumption data with appropriate dimensionality reduction technique such that it captures the relations present in the data without loss of important information and low training time? Moreover, (iv) How to leverage the power of generative models for building a robust electricity theft detection model that can provide a high detection rate and less false alarm rate, especially for unseen data? The solution to the above research questions highlighted as the significant contributions are:
\begin{itemize}
  \item EnsembleNTLDetect, a robust and scalable framework for detecting NTLs in smart grids through analysing consumption patterns from the real-time energy consumption data, is presented.
    \item The efficiency of enhanced Dynamic Time Warping based Imputation (eDTWBI) is improved by introducing a $Search\_Size$ parameter to reduce the search space of eDTWBI and thereby provides an effective way to handle the large missing gaps in the time series data. 
    \item  A customised stacked autoencoder is designed to handle the high dimensional electricity consumption data. The 1,034 dimensions in the original dataset were reduced to 128 dimensions while retaining 99.87\% of the original data with reduced training time.
    \item Conditional GAN is fine-tuned to aid the robust training of classifiers. During the training phase, the classifiers are exposed to real and synthetic data so that the classifiers can model different types of energy consumption values accurately with high confidence scores.
    \item A soft voting ensemble classifier was designed to leverage the combined efficiency of the bagging-boosting technique based on Random Forest and XGBoost algorithms to achieve a high detection rate and low false alarm rate.
    \item EnsembleNTLDetect is validated using the real-time electricity consumption data obtained from State Grid Corporation of China (SGCC) using various quality metrics.
\end{itemize}
The paper is structured as follows: Section \ref{Methodology} provides a detailed insight into the architecture and workflow of EnsembleNTLDetect. Section \ref{result} highlights the experimental analysis carried out to demonstrate the performance of EnsembleNTLDetect over the state-of-the-art electricity theft detection models in terms of various quality metrics, and Section \ref{conclusion} concludes the paper with the scope for further research.

\begin{figure}[t]
\centering
\begin{tabular}{c}
    \includegraphics[width=80mm]{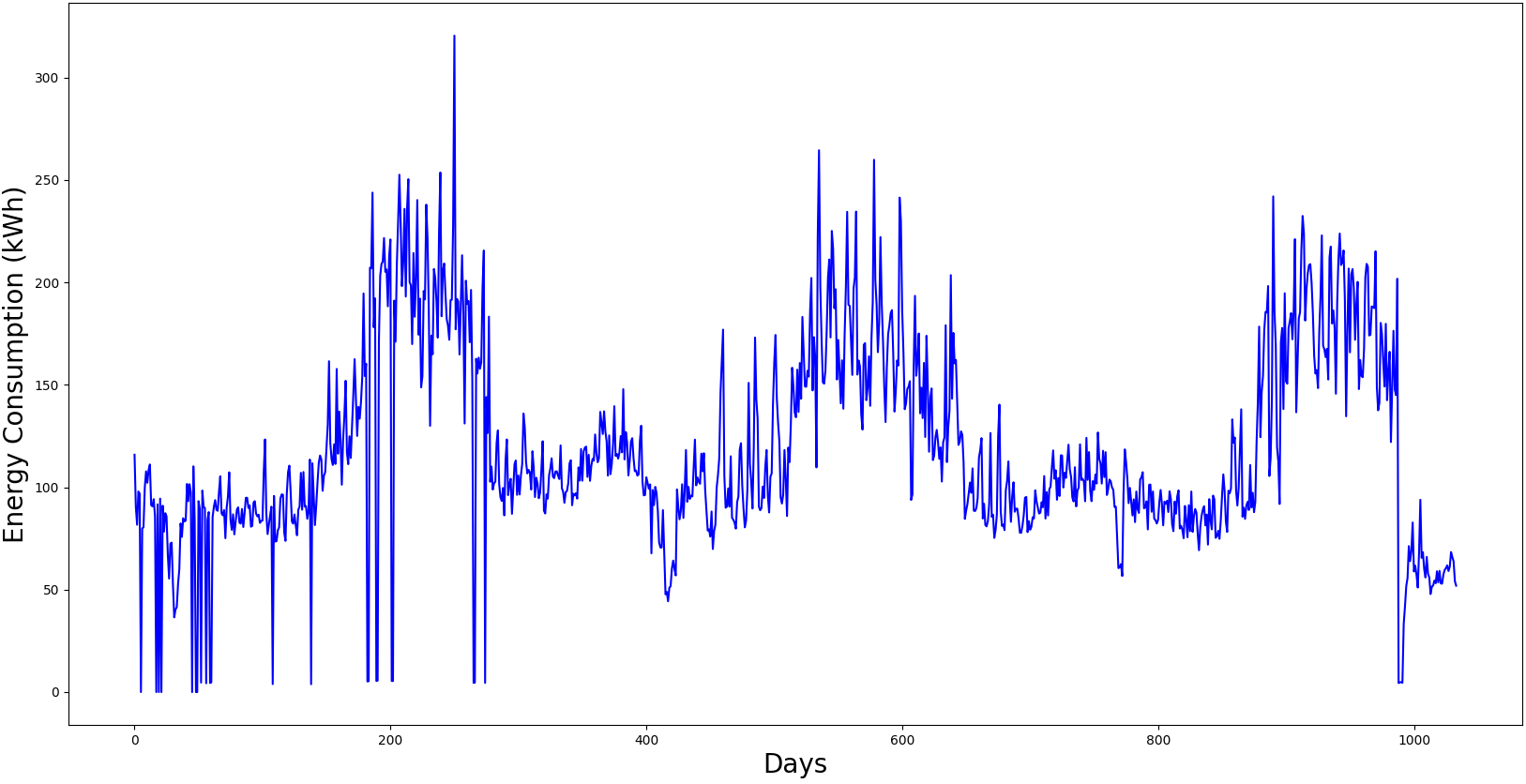} \\
      (a) Genuine consumer\\
    \includegraphics[width=80mm]{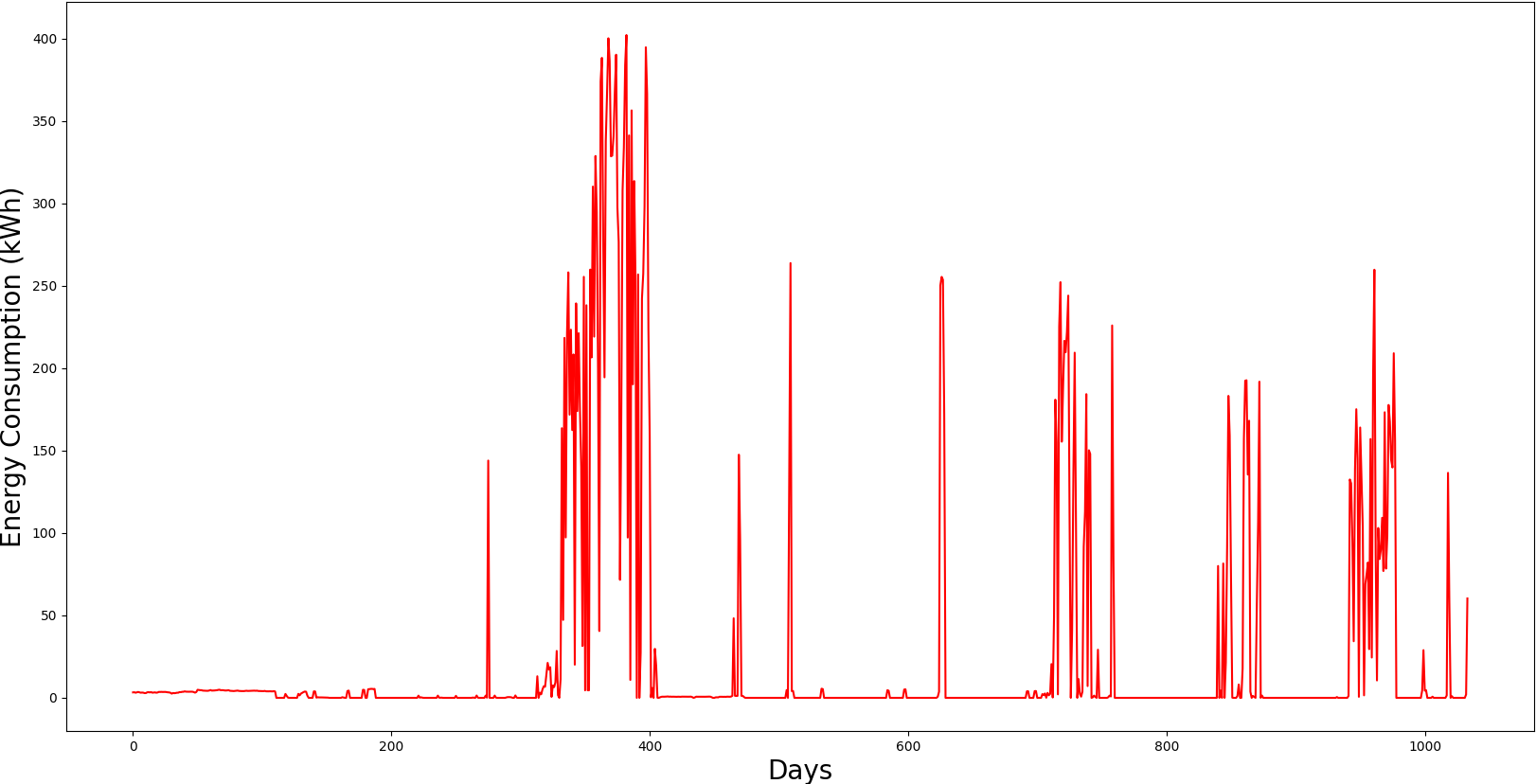} \\
  (b) Electricity thieve\\
\end{tabular}
\caption{Electricity consumption patterns of genuine user and electricity thieve}
\label{comparison_normal_abnormal}
\end{figure}

\section{Methodology}\label{Methodology}
Figure \ref{complete_block} presents the overall architecture of the EnsembleNTLDetect, the proposed electricity theft detection model. The complete working methodology of EnsembleNTLDetect with five stages, namely (i) data acquisition and preprocessing, (ii) data sampling, (iii) learning model, (iv) model tuning and training, and (v) evaluation, for efficient and reliable energy theft detection, is detailed below.

With the scarcity of open-source electricity consumption data, this study uses a real-time electricity consumption dataset released by State Grid Corporation of China (SGCC) \cite{b32}. The SGCC dataset comprises of daily electricity consumption of 42,372 consumers with 38,757 genuine consumers (\textit{class 0}) and 3615 electricity thieves (\textit{class 1}) recorded over a period of 2 years (1st January 2014 to 31st October 2016). A closer observation to Figure \ref{comparison_normal_abnormal} states that the electricity consumption pattern of electricity thieves is aberrant (with more spikes and low) than the genuine consumers. 
In general, electricity consumption data recorded from the smart meters is aggregated and transferred over data channels to a central location for storage and processing. However, as a result of sensor failures, transmission errors, and server issues, the major challenges in the application of the SGCC time series dataset for electricity theft detection is three-fold (i) 11,233,528 missing values, (ii) imbalanced data in the ratio of 10:1, and (iii) outliers.

\subsection{Data Preprocessing}
\subsubsection{\textbf{Missing value imputation}}

The SGCC dataset contains about 11,233,528 missing values which approximates about 25\% of the dataset. Ignoring such missing values might lead to downsizing the dataset, which poses a significant challenge in carrying out reliable analysis. Previous works \cite{b11, b23, b24, b25} have used linear interpolation, mean of previous and following day consumption's, filling with mean or median of a complete column, and dropping rows which have missing values beyond a certain threshold. Such methods perform well for isolated data, i.e., one to three missing values but fail miserably for realistic imputations in data with consecutive missing values, correlations, seasonality trends and complex distribution.

\begin{figure}[t]
\centering
\begin{tabular}{c}
    \includegraphics[width=80mm]{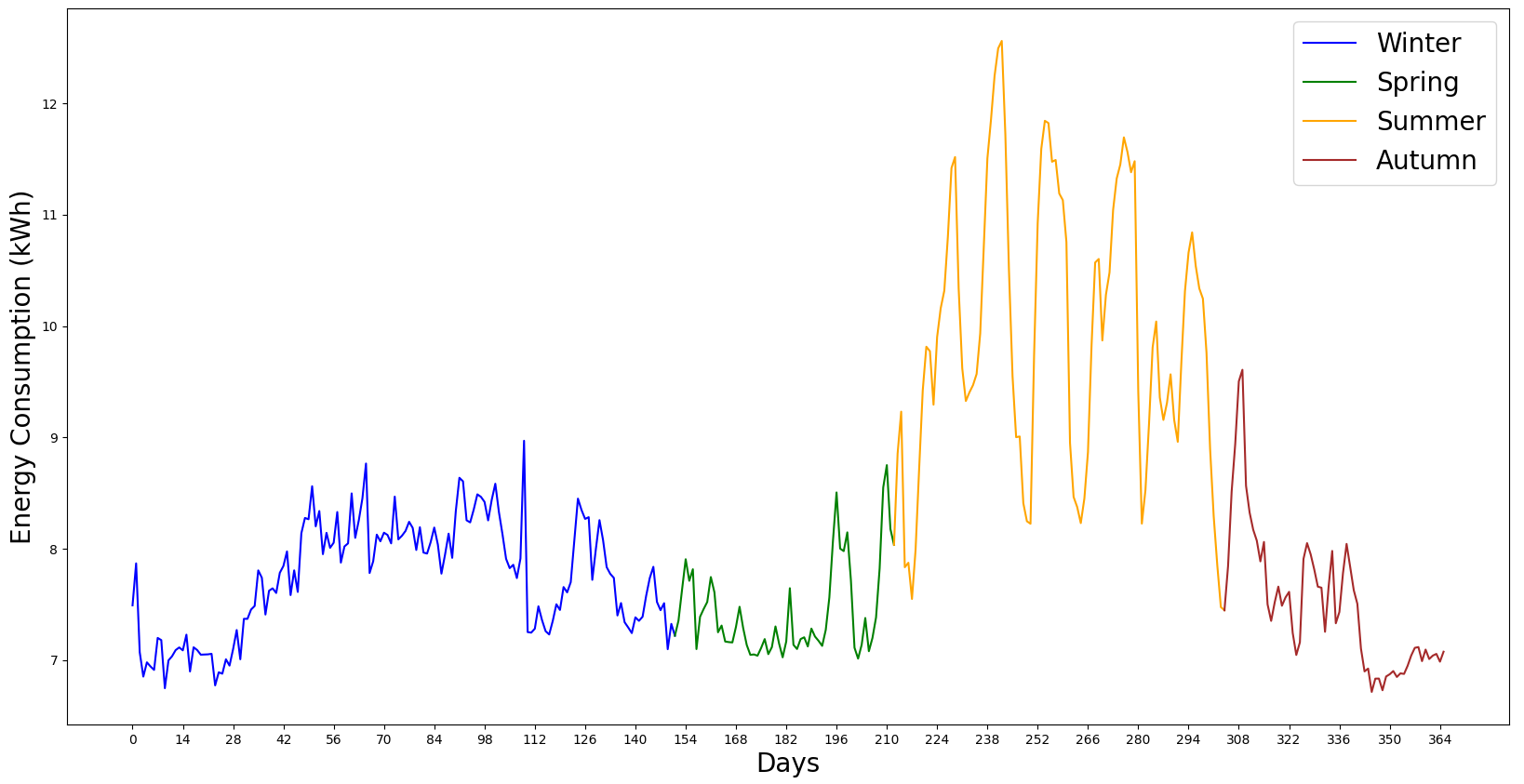}\\
\end{tabular}
\caption{Variations in electricity consumption over different seasons}
\label{Seasonal_plot}
\end{figure}

\textbf{RQ1:} \textit{How to handle large gaps, i.e., consecutive missing values, in time series data with high seasonal trends in an effective way?}\\ This work employs an enhanced version of Dynamic Time Warping (DTW) based Imputation (eDTWBI) \cite{b33}, an algorithm for the generation of optimal time series data, i.e., to fill the large gaps (consecutive missing values) in the SGCC dataset. eDTWBI uses DTW to find two reference window that lies before and after the gap, which is also similar to the considered large gap such that the distance between them is minimal. The reference window is represented as grids for quadratic time complexity. Owing to the large size of the SGCC dataset, this work introduces \textit{\text{Search\_Size}} parameter to reduce the search space. Further, the seasonality trends in the dataset are taken care of by \textit{\text{Search\_Size}} to ensure that the imputation for gaps in a particular season is bounded by the similar sequences obtained from the same season. For example, gaps in the summer season ($Search\_Size$ = 1) are imputed using the similar sub-sequence obtained within the year's summer season (3 months). Figure \ref{Seasonal_plot} shows the seasonality trend of the dataset for the year 2015. This significant improvement in eDTWBI helps enhance the learning base, prediction ability, data dynamics and reduces the temporal constraints between the reference window. Algorithm \ref{alg1} presents the pseudo-code of the eDTWBI algorithm for missing value imputation.

\begin{algorithm}[t]
\SetAlgoLined
\KwInput{$x=\left\{x_{1}, x_{2}, \ldots, x_{N}\right\}$, $t$, $T$, \(Search\_Size = 0\), $Q = D[t - T : t -1]$, $lp = $ [ ]}
  \KwOutput{Imputed DataFrame}
  Construct a DTW\_Matrix $D$ consisting of $n$ rows and $m$ columns where $(m,n)$ $\in$ $len(sequence)$ and $D_{ij} = distance(x_{i}, x_{j})$\\ 
  Create a $Search\_Space$ $S = D[1 : t - 2T]$ \\
  Set $Derivative\_Cost\_Measure$ for DTW algorithm using the following formula:
 \begin{equation}
    D_{x}[a]=\frac{\left(x_{a}-x_{a-1}\right)+\left(\left(x_{a+1}-x_{a-1}\right) / 2\right)}{2}
\end{equation}
\begin{equation}
    Derivative\_Cost = (D_{x}[i] - D_{x}[j])^2
\end{equation}\\
$i \leftarrow 1$ \&  $Search\_Size \leftarrow 3$\\
\While{$ i < len(S)$}
{
    $k \leftarrow i+T-1$\\
        Save a reference window $R_{Before}(i) = S[i : k]$\\
        \eIf{$ R_{Before}(i)$ in $Search\_Size$}
        {
            $dtw\_cost = DTW(Q, R_{Before}(i))$\\
             \eIf{$dtw\_cost < Derivative\_Cost$}
            {
                $i \leftarrow i+1$
                
            }
            {
               Save position of $R_{Before}(i)$ to $lp$
            }
        }
        {
        break
        }
}
Replace all missing values at position $t$ by an array of values after the $Q$'s window having minimum $DTW$ cost using the $lp$ list.\\
\Return{Imputed Dataframe}
\caption{Enhanced Dynamic Time Warping with Reduced Search Space}
\label{alg1}
\end{algorithm}

\begin{figure}[t]
\centering
\begin{tabular}{c}
    \includegraphics[width=80mm]{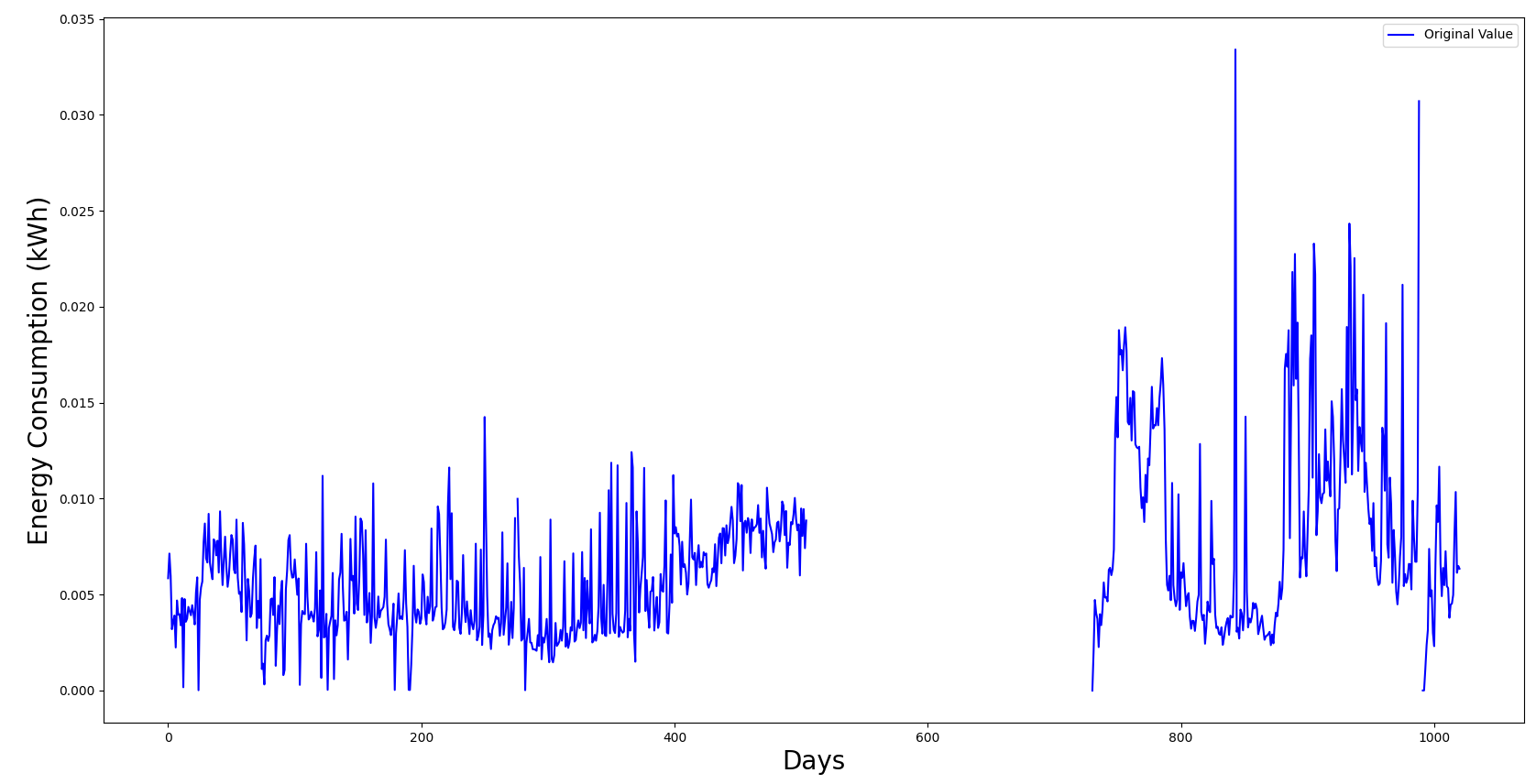} \\
      (a) Consumption patterns with large gap \\
    \includegraphics[width=80mm]{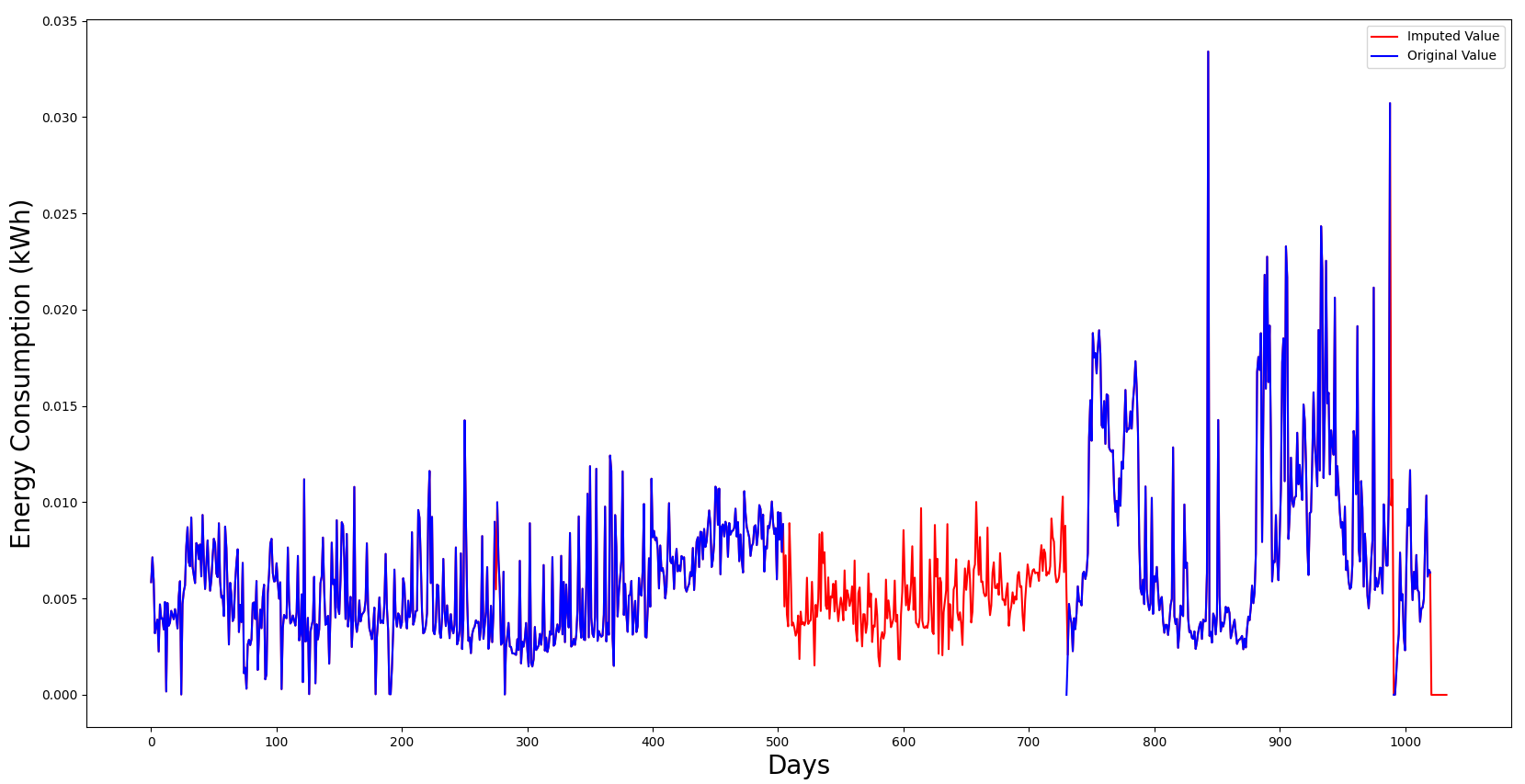} \\
  (b) Consumption patterns with imputed values\\
\end{tabular}
\caption{Imputation using Algorithm \ref{alg1}}
\label{imputed}
\end{figure}

For a time-series $x$ with large number of consecutive missing values, a gap of size \(T\) at position \(t\) is defined as the portion between two points \(x_t\) and \(x_{t+T-1}\) that has \(x_i\) \textit{NaN} values, where \(i = t : t + T -1 \). Further, $Q$ forms the temporal window before missing values, $R$ is a reference window for imputation that should lie within the same season, and $lp$ is an array of location pointers pointing to reference windows with a minimum $DTW$ cost. The workflow of eDTWBI is highlighted below:
\begin{enumerate}[Step 1:]
\item Create reference window: For a gap of \(T\) size at position \(t\), create two reference windows ($R_{Before}$ \& $R_{After}$) containing data points that lie before and after the gap of length \(T\). 
\item Find highly similar windows: Create sliding windows of length \(T\) for the data points that lie before and after the gap \(T\). Identify the most similar window to reference windows ($R_{Before}$ \& $R_{After}$) by calculating the $DTW$ cost and Derivative Dynamic Time Warping ($DDTW$) cost \cite{b34}. Since $DDTW$ cost is robust to outliers, save the windows whose $DDTW$ cost is lesser than the $DTW$ cost.
\item Imputation: For unbiased results, impute the large gap of length \(T\)) with the average value of the most similar windows. 
\end{enumerate}

Figure \ref{imputed} provides the electricity consumption data with large gaps and imputed values. The complete process of imputation for the whole dataset took less than \textit{30 minutes}. The novelty introduced in the eDTWBI algorithm in terms of restricting the search space has improved the efficiency of the EnsembleNTLDetect in terms of performance measures and execution time.

\subsubsection{\textbf{Outliers}}
The imputed dataset is subjected to outlier detection and removal using \textit{Z-score}, a computationally inexpensive outlier removal technique given by equation \ref{z_eqn}. 
\begin{equation}\label{z_eqn}
Z=\frac{\mathrm{X}-\mu}{\sigma}
\end{equation}
Where $X$ refers to the data point, $\mu$ is the mean, $\sigma$ is the standard deviation, and $Z$ is the \textit{Z-score}.
All data points which have $Z > 3$ or $Z < -3$ were dropped. 

\subsection{Handling Imbalanced Data}
\textbf{RQ 2:} \textit{What is the impact of undersampling techniques on the generation of balanced electricity consumption dataset without information loss?}\\The SGCC dataset is imbalanced in the ratio of 10:1 with class 0 (genuine consumers) as the majority and class 1 (electricity thieves) as the minority. Although SMOTE has been widely used in the literature to handle data imbalance issues in the SGCC dataset, this work prefers to use the under-sampling technique due to the following reasons. 
\begin{itemize}
    \item From figure \ref{comparison_normal_abnormal}, it is clear that the electricity consumption pattern of electricity thieves is aberrant when compared with the genuine consumer’s consumption trend. In general, the application of SMOTE generates such unusual patterns with unrealistic consumption values for the minority class (electricity thief) and are highly susceptible to overfitting. Such unusual patterns pose difficulty for the classifiers in extracting meaningful information and accurate classification; furthermore, increasing the number of samples in the minority class results in low accuracy and a minute increase in recall score. In such cases, the application of under-sampling approaches ensures the classifier to establish a fine boundary between genuine consumers with usual trend and electricity thieve with unusual consumption trends. Refer Section \ref{result} for more details.
\item The computational efficiency of under-sampling techniques is another reason for its consideration over SMOTE. 
\end{itemize}

\begin{table}[t]
\caption{Performance comparison of Random Forest using SMOTE \& Near Miss}
\label{smote_nearmiss_table}
    \centering
    \begin{tabular}{|l|l|l|l|}
    \hline
    \textbf{Output} & \textbf{Parameter} & \textbf{SMOTE $+$ RF} & \textbf{Near--miss $+$ RF} \\
    \hline
    0 (genuine) & Precision & 0.94 & 0.98\\
    \hline
    1 (theft) & Precision & 0.39 & 0.62\\
    \hline
    0 & Recall & 0.96 & 0.99 \\
    \hline
    1 & Recall & 0.29 & 0.57 \\
    \hline
    0 & F1-score & 0.95 & 0.98\\
    \hline
    1 & F1-score & 0.33 & 0.59 \\
    \hline
    \end{tabular}
\end{table}

\begin{table*}[t]
\tiny
\caption{Architecture of Stacked Auto-Encoder consisting of three Auto-Encoder's}
\label{architecture_table}
\centering
\begin{tabular}{ |c|c|c|c|c|c|c|c| }
\hline
\multicolumn{2}{|c|}{Stacked Auto-Encoder} & \multicolumn{2}{|c|}{Auto-Encoder 1} & \multicolumn{2}{|c|}{Auto-Encoder 2 } & \multicolumn{2}{|c|}{Auto-Encoder 3}\\
\hline
Layers & Parameters & Layers & Parameters & Layers & Parameters & Layers & Parameters\\
\hline
Input & (1034, ) & Input & (1034, ) & Input & (512, ) & Input & (256, )\\
Dense & 512, ReLU, param = 529,920 & Dense & 512, ReLU, param = 529,920 & Dense & 256, ReLU, param = 131,328 & Dense & 128, ReLU, param = 32,896\\
Batch-Norm & param = 2,048 & Batch-Norm & param = 2,048 & Batch-Norm & param = 1,024 & Batch-Norm & param = 512\\
Dense & 256, ReLU, param = 131,328 & Dense & 1034, Sigmoid, param = 530,442 & Dense & 512, Sigmoid, param = 131,584 & Dense & 256, Sigmoid, param = 33,024\\
Batch-Norm & param = 1,024 & & & Batch-Norm & param = 2,048 & Batch-Norm & param = 1,024\\
Dense & 128, ReLU, param = 32,896 & & & & & &\\
Batch-Norm & param = 512 & & & & & &\\
Dense & 256, Sigmoid, param = 33,024 & & & & & &\\
Batch-Norm & param = 1,024 & & & & & &\\
Dense & 512, Sigmoid, param = 131,584 & & & & & &\\
Batch-Norm & param = 2,048 & & & & & &\\
Dense & 1034, Sigmoid, param = 530,442 & & & & & &\\
\hline
Total Parameters & 1,395,850 & Total Parameters & 1,062,410 & Total Parameters & 265,984 & Total Parameters & 67,456 \\
\hline

\end{tabular}
\end{table*}

This work employs \textit{Near-Miss} (version 1) \cite{b35}, a simple and effective under-sampling technique to handle the data imbalance in the SGCC dataset. About 40,488 samples obtained after applying the z-score outlier removal technique were reduced to 6,300 with an equal split of 3,150 samples for class 0 and class 1. An interesting point to note is that the \textit{meaningful} information lost, if any, during under-sampling was handled by the CTGAN. 

We chose random forest classifier since it is part of our proposed soft voting ensemble for performing this experiment. From Table \ref{smote_nearmiss_table} we can infer that for the SGCC dataset, near-miss undersampling works better in comparison with SMOTE oversampling. The recall score for SMOTE (theft class) is 0.29 since it augments the aberrant minority samples making the task more challenging for the classifier. In contrast, it is 0.57 for Near Miss since it downsamples the majority class having certain periodicity.

\subsection{Learning Model}
\subsubsection{\textbf{Stacked AutoEncoder}}
With 1,034 timestamps in the SGCC dataset, it is evident that these features carry some intrinsic relation between them. Therefore, applying the dimensionality reduction technique to identify a set of informative features is the ideal step towards the design of an efficient and reliable electricity theft detection model. Unfortunately, principal component analysis \cite{b36, b37}, the most commonly used dimensionality reduction technique, fails to capture the convoluted low-dimensional manifold structure and model the intrinsic relations in the time series data \cite{b38}.

\begin{figure}[b]
\centering
\begin{tabular}{c}
    \includegraphics[width=80mm]{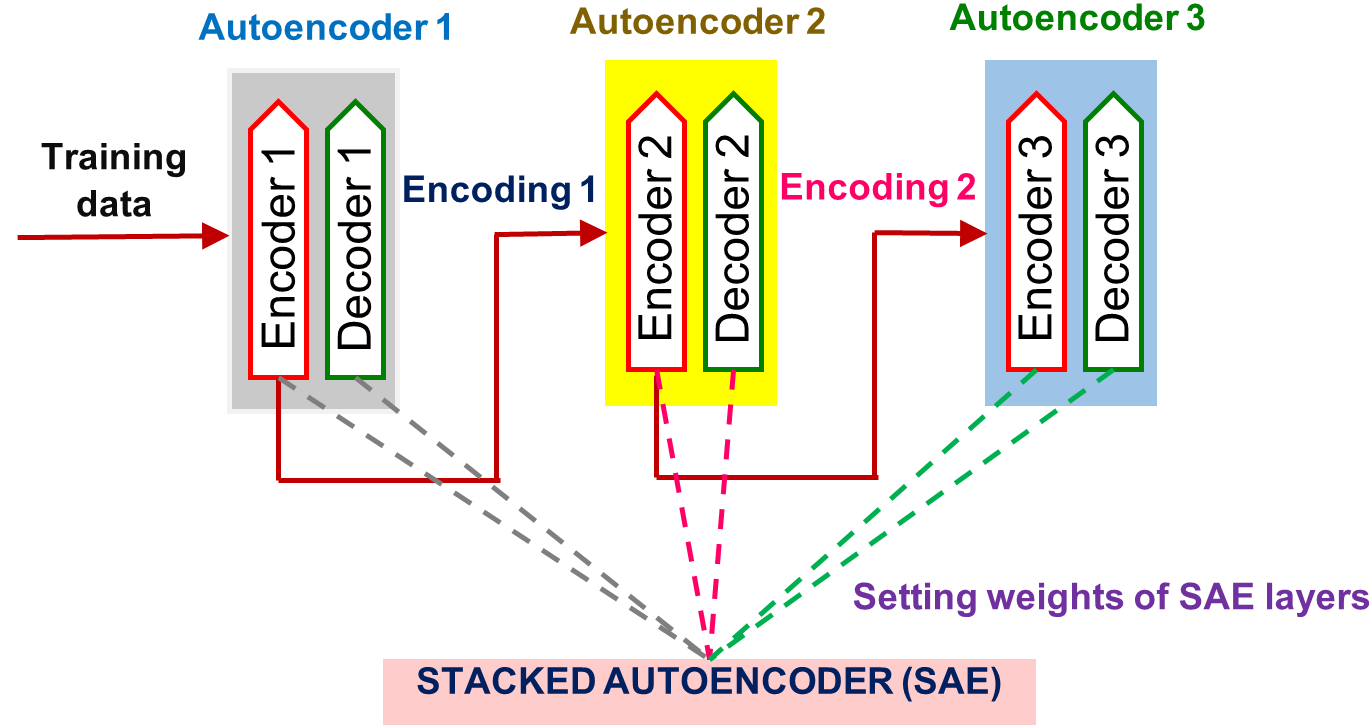}\\
\end{tabular}
\caption{Autoencoder training pipeline}
\label{autoencoder_block}
\end{figure}

\begin{figure}[ht]
\begin{minipage}{.5\linewidth}
\centering
\subfloat[AE1]{\includegraphics[width=40mm]{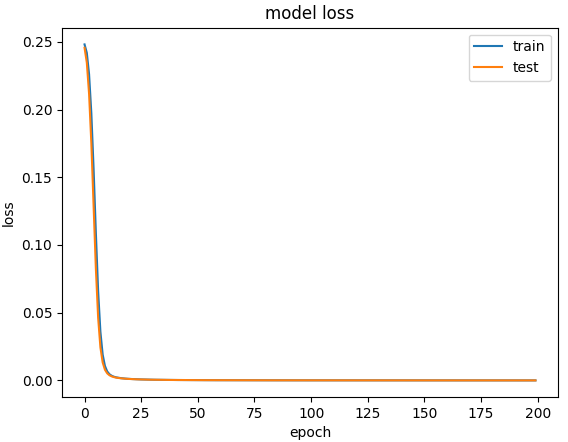}}
\end{minipage}%
\begin{minipage}{.5\linewidth}
\centering
\subfloat[AE2]{\includegraphics[width=40mm]{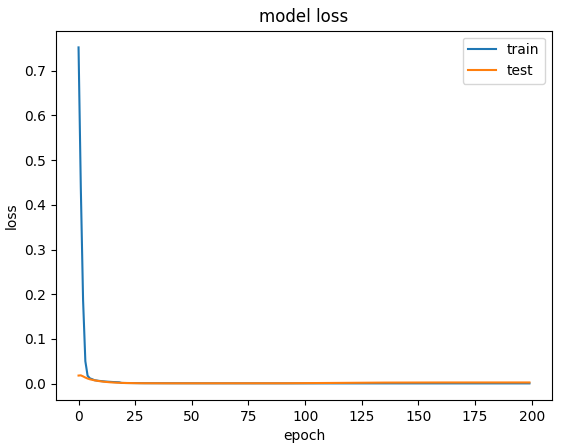}}
\end{minipage}\par\medskip
\centering
\subfloat[AE3]{\includegraphics[width=40mm]{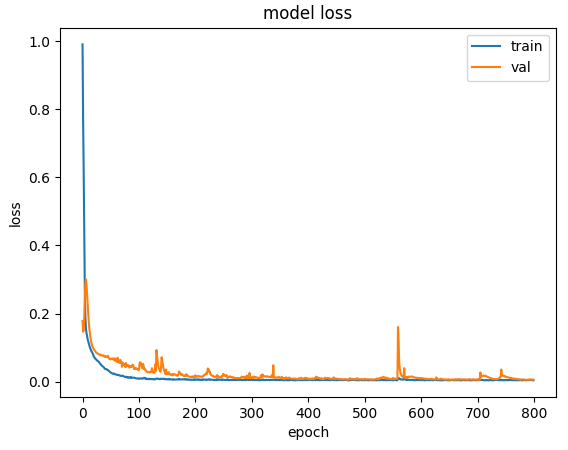}}
\caption{Model Loss during training for the AE's}
\label{model_loss}
\end{figure}

\textbf{RQ 3:} \textit{How to handle the high dimensional electricity consumption data with appropriate dimensionality reduction technique such that it captures the relations present in the data without loss of important information and low training time?}\\
In such cases, autoencoder architectures have been successfully established as an efficient dimensionality reduction tool for fault diagnosis \cite{b39}, high-content screening data \cite{b40} and intrusion detection systems \cite{b41}. Autoencoders are a special kind of neural networks which maps the input of a specific dimension to a latent space of reduced dimension and then decode the latent representation to a reconstructed input by minimizing the reconstruction error. This work presents a stacked autoencoder with three autoencoders specifically designed to perform an unsupervised learning based dimensionality reduction on the feature space. Table \ref{architecture_table} shows the model architecture of the stacked autoencoder with three autoencoders. The weights of the hidden layers are set by training each autoencoder individually. Figure \ref{autoencoder_block} presents the training procedure of the autoencoder for dimensionality reduction, and figure \ref{model_loss} presents the model loss (during training) for the three autoencoders, we can see that our model converges quickly within 100 epochs. The proposed stacked autoencoder has reduced 1,034 dimensions in the SGCC dataset to 128 dimensions, wherein  99.87\% of the original data was captured with no loss of information. The application of autoncoder based dimensionality reduction technique has boosted the efficiency of this framework resulting in faster training and inference.

\textbf{RQ 4:} \textit{How to leverage the power of generative models for building a robust electricity theft detection model that can provide a high detection rate and less false alarm rate, especially for unseen data?}

\begin{table*}[t]
\caption{Performance analysis of EnsembleNTLDetect and basic machine learning architectures for NTL detection}
\label{basic_table}
\centering
\begin{tabular}{ |l|l|l|l|l|l|l|}
\hline
\textbf{Classifier} & \textbf{Precision} & \textbf{Recall} & \textbf{F1-Score} & \textbf{AUC-ROC} & \textbf{PR-AUC} & \textbf{MCC}\\
\hline
Naive Bayes & 0.65 & 0.53 & 0.54 & 0.54 & 0.12 & 0.14\\
\hline
ExtraTrees & 0.56 & 0.56 & 0.56 & 0.58 & 0.11 & 0.12\\
\hline
K--Neighbors & 0.65 & 0.52 & 0.53 & 0.64 & 0.16 & 0.12\\
\hline
Linear Support Vector Machine & 0.78 & 0.50 & 0.47 & 0.67 & 0.22 & 0.03\\
\hline
Logistic Regression & 0.75 & 0.50 & 0.49 & 0.68 & 0.21 & 0.09\\
\hline
Multi--layer Perceptron & 0.73 & 0.56 & 0.58 & 0.77 & 0.32 & 0.24\\
\hline
Random Forest & 0.79 & 0.53 & 0.54 & 0.80 & 0.34 & 0.21\\
\hline
Gradient Boosting & 0.76 & 0.54 & 0.55 & 0.79 & 0.33 & 0.21\\
\hline
\textbf{EnsembleNTLDetect} & \textbf{1.00} & \textbf{0.98} & \textbf{0.99} & \textbf{0.99} & \textbf{0.99} & \textbf{0.98}\\
\hline
\end{tabular}
\end{table*}

\subsubsection{\textbf{Handling Corner Cases}}
\begin{figure}[b]
\centering
\begin{tabular}{c}
    \includegraphics[width=80mm]{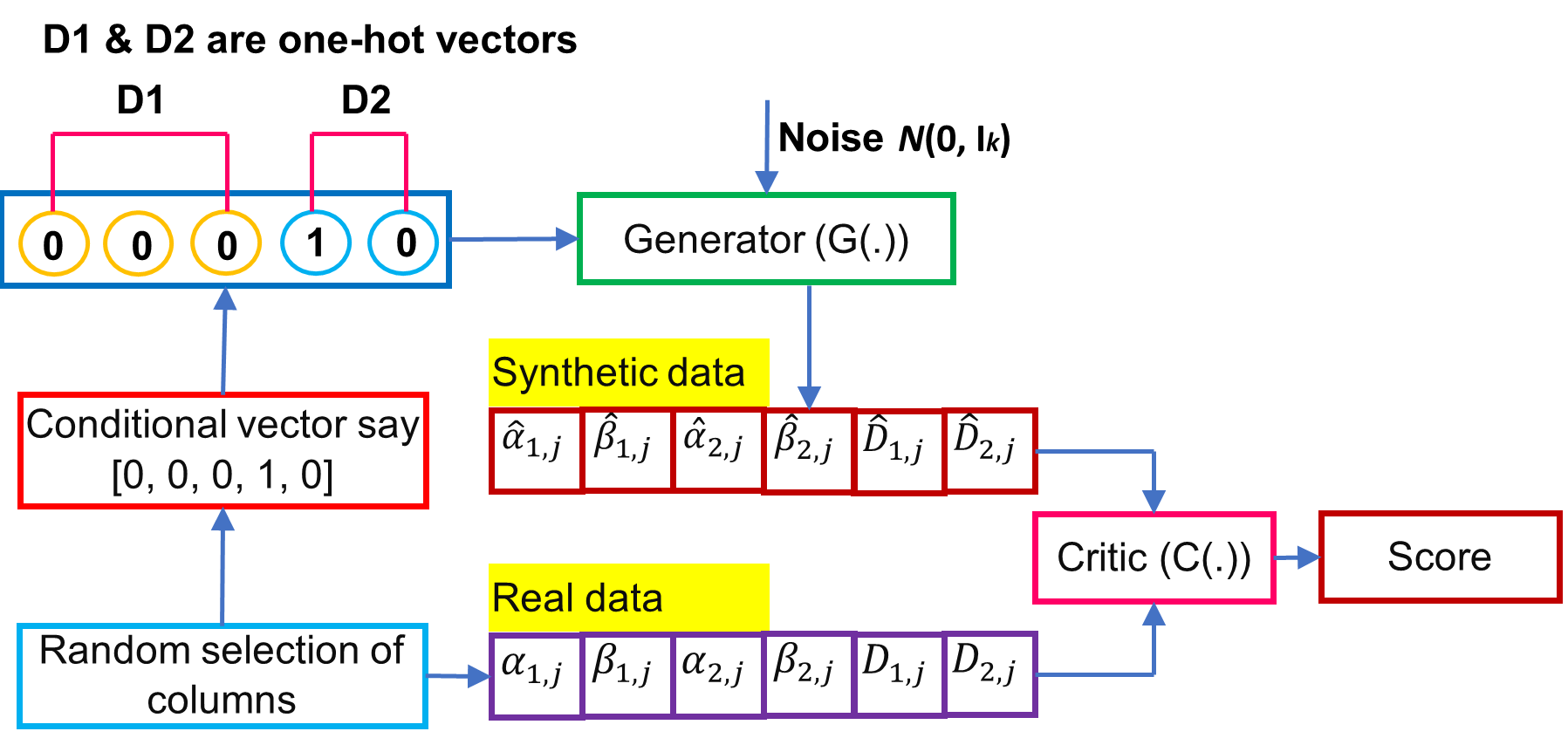}\\
\end{tabular}
\caption{Architecture of CTGAN}
\label{ctgan_block}
\end{figure}

\begin{table}[b]
\caption{Optimal Hyper-Parameters} 
\centering 
\begin{tabular}{| l | l |} 
\hline
\label{hyperparameter_table}
\textbf{Algorithm} & \textbf{Parameters} \\
\hline
Random Forest & \makecell[l]{n\_estimators=300, max\_features="sqrt",\\ criterion="gini", min\_samples\_leaf=5,\\ class\_weight="balanced"} \\
\hline
XGBoost & \makecell[l]{objective="binary:logistic", learning\_rate=0.03,\\ n\_estimators=500, max\_depth=1,\\ subsample=0.4} \\
\hline
Logistic Regression & penalty="l2",C=100\\
\hline
\end{tabular}
\end{table}

To ensure the efficiency and reliability of EnsembleNTLDetect in real-time environments, it is highly essential to look on to the critical aspects such as loss of critical information due to undersampling and ability to handle various input types. To handle such issues, this work uses Conditional Tabular Generative Adversarial Network (CTGAN) \cite{b42} to create more samples in such a way that the learning model is exposed to a wide range of data samples from both the classes (genuine and electricity thieves). Figure \ref{ctgan_block} shows the architecture of CTGAN. 

CTGAN improves the tabular data generation through (i) mode-specific normalization: improves modelling multi-modal distributions in numeric columns and (ii) conditional training by sampling: ensures that the rare categorical data are evenly sampled. Since the SGCC dataset comprises of daily electricity consumption values (numerical), mode-specific normalization was preferred over the conditional training. The number of modes for each column determined by the Variational Gaussian Mixture (VGM) \cite{b43} was used for normalization. These normalized values were used during the training phase and are transformed to their original scales after obtaining the generated data. Due to the complexity involved in training GANs, Wasserstein GAN with gradient penalty \cite{b44} and PacGAN \cite{b45} were used to ensure robust learning stability (prevents mode collapse), and the generator network provides diverse samples, respectively. For the SGCC dataset, 10,000 samples in the ratio of 2:1 between genuine consumers and electricity thieves were generated using CTGAN.

\subsubsection{\textbf{Soft voting ensemble classifier}}
A simple soft voting classifier with Random Forest \cite{b46}, and XGBoost \cite{b47} classifiers as base learners were designed for accurate classification of genuine consumers and electricity thieves with high detection rates and less false alarm rate. Further, logistic regression was used as a meta learner to create a linear relationship between the input and output variables, i.e., a fine boundary between genuine consumers and electricity thieves using the maximum-likelihood estimation based on coefficients obtained from the training data. A soft voting mechanism is used so that the output class has the highest average probability. The output label $\hat{y}$ of a soft voting ensemble model with $m$ classifiers of $p$ probability is given in equation \ref{vote_eqn} where, $w_j$ is the uniform weight of the $j^{th}$ classifier, $i \in\{0,1\}$. 
\begin{equation}\label{vote_eqn}
\hat{y}=\arg \max _{i} \sum_{j=1}^{m} w_{j} p_{i j}
\end{equation}

The optimal hyperparameters of the ensemble classifier were obtained through rigorous 10-fold cross-validation using GridSearchCV with the best validation accuracy. Table \ref{hyperparameter_table} presents the optimal hyperparameters of the ensemble classifier.

\begin{figure}[t]
\centering
\begin{tabular}{l}
    \includegraphics[width=80mm]{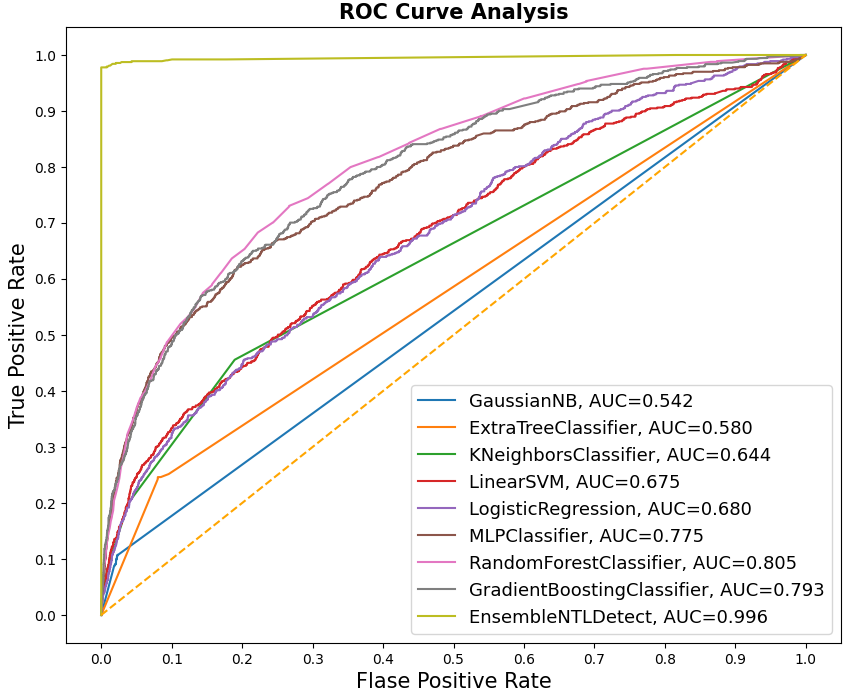}\\
\end{tabular}
\caption{ROC Curve of EnsembleNTLDetect and basic ML architectures}
\label{normal_auc_roc}
\end{figure}

\section{Results \& Discussions}\label{result}
The experimental design and analysis of EnsembleNTLDetect was carried out in a working environment with an Intel i5 (10th Gen) processor running Windows 10 operating system with 8 GB RAM. For faster implementations, EnsembleNTLDetect and the comparative models were designed and executed in Google Colaboratory. Moreover, the implementation of the EnsembleNTLDetect and the comparative models was done using Python 3.7 with necessary packages and libraries such as scikit-learn, TensorFlow, Keras.
\subsection{Performance Metrics}\label{Performance Metrics}
Due to the imbalanced nature of the SGCC electricity consumption dataset, the essential quality metrics derived from the confusion matrix were chosen over accuracy to assess the performance of EnsembleNTLDetect over the state-of-the-art electricity theft detection models. Here, the primary measures of the confusion matrix represent (i) True Positive ($TP$): correctly classified as electricity thieve, (ii) True Negative ($TN$): correctly classified as genuine consumer, (iii) False Positive ($FP$): misclassified as electricity thieve and (iv) False Negative ($FN$): misclassified as a genuine consumer.

\begin{table*}[t]
\caption{Performance analysis of EnsembleNTLDetect and state-of-the-art electricity theft detection models}
\label{cmp_table}
\centering
\begin{threeparttable}
\begin{tabular}{ |l|l|l|l|l|l|l|}
\hline
\textbf{Classifier} & \textbf{Precision} & \textbf{Recall} & \textbf{F1-Score} & \textbf{AUC-ROC} & \textbf{PR-AUC} & \textbf{MCC}\\
\hline
SVM \cite{b11} & 0.75 & 0.71 & 0.72 & 0.60 & 0.78 & 0.67\\
\hline
XGBOOST \cite{b48} & 0.95 & 0.82 & 0.86 & 0.88 & 0.87 & 0.81\\
\hline
Bi-directional Gated Recurrent Unit \cite{b49} & 0.82 & 0.82 & 0.84 & 0.84 & 0.78 & 0.68\\
\hline
CNN + RF \cite{b23} & 0.80 & 0.89 & 0.85 & 0.90 & 0.87 & 0.84\\
\hline
Wide CNN \cite{b24} & 0.84 & 0.88 & 0.86 & 0.86 & 0.81 & 0.73\\
\hline
CNN + LSTM \cite{b25} & 0.94 & 0.82 & 0.88 & 0.88 & 0.87 & 0.78\\
\hline
LSTM + MLP \cite{b50} & 0.90 & 0.87 & 0.85 & 0.90 & 0.90 & 0.80\\
\hline
Semi-Supervised AutoEncoder \cite{b51} & 0.86 & 0.80 & 0.83 & 0.84 & 0.81 & 0.82\\
\hline
\textbf{EnsembleNTLDetect} & \textbf{1.00} & \textbf{0.98} & \textbf{0.99} & \textbf{0.99} & \textbf{0.99} & \textbf{0.98}\\
\hline
\end{tabular}
\begin{tablenotes}
    \item All the comparison with the previous work was done keeping test size = 0.2.
\end{tablenotes}
\end{threeparttable}
\end{table*}

\subsubsection{Precision}
The ratio of consumers (thieves) correctly classified as electricity thieves to the total positive predictions.
\begin{equation}\label{precision}
\text { Precision }= \frac{T P}{T P+F P}
\end{equation}
\subsubsection{Recall / True Positive Rate (TPR)} The ratio of consumers (thieves) correctly classified as thieves to all the predictions of actual class.
\begin{equation}\label{recall}
\text { Recall }= \frac{T P}{T P+F N}
\end{equation}
\subsubsection{F1-score} The harmonic mean of precision and recall.
\begin{equation}\label{f1}
F1-\text { Score }= 2 * \frac{\text { Precision } * \text { Recall }}{\text { Precision }+\text { Recall }}
\end{equation}
\subsubsection{AUC-ROC} A probability curve that plots the TPR against FPR.
\begin{equation}
\text { FPR }= \frac{F P}{T N+F P}
\end{equation}
\subsubsection{PR-AUC} Represents the precision against recall score over varying thresholds. A high score indicates that a classifier can accurately achieve $TPs$ with very less number of $FPs$ \& $FNs$.
\subsubsection{Matthews Correlation Coefficient (MCC)} The most reliable statistical measure for imbalanced data. A high score represents that the classifier performed well for all categories of the confusion matrix ($TPs$, $FPs$, $TNs$, $FNs$).
\begin{equation}\label{mcc}
\tiny
\mathrm{MCC}=\frac{T P \times T N-F P \times F N}{\sqrt{(T P+F P)(T P+F N)(T N+F P)(T N+F N)}}
\end{equation}

\subsection{Analysis and discussions}
At the initial phase, the performance of the EnsembleNTLDetect was evaluated over the basic machine learning architectures like in terms of the quality metrics mentioned in Section \ref{Performance Metrics}. The metric values provided in this section are the average values obtained after 25 consecutive and iterative runs. Table \ref{basic_table} presents a comparative analysis of EnsembleNTLDetect over the basic machine learning architectures for NTL detection. Even though it is evident that EnsembleNTLDetect demonstrates its performance with better quality metrics, the contrast models can be categorized into two groups, (i) Type 1: Naive Bayes, Extra Trees, K-nearest neighbours, Linear SVM and logistic regression (AUC-ROC $<$ 0.75 and MCC $<$ 0.15) and (ii) Type 2: MLP, random forest and gradient boosting (AUC-ROC $<$ 0.85 and MCC $<$ 0.25) based on the AUC-ROC and MCC scores. Figure \ref{normal_auc_roc} provides the AUC-ROC curve of EnsembleNTLDetect and basic machine learning-based NTL detection models.

\begin{table}[b]
\caption{Performance analysis of EnsembleNTLDetect for different train-test splits}
\label{split_table}
    \centering
    \begin{tabular}{|l|l|l|l|l|l|}
    \hline
    \textbf{Split Size} & \textbf{Precision} & \textbf{Recall} & \textbf{F1-Score} & \textbf{AUC-ROC} & \textbf{MCC} \\
    \hline
    70 : 30 & 0.995 & 0.978 & 0.987 & 0.994 & 0.974\\
    \hline
    60 : 40 & 0.995 & 0.979 & 0.987 & 0.993 & 0.975\\
    \hline
    50 : 50 & 0.998 & 0.972 & 0.985 & 0.989 & 0.971\\
    \hline
    \end{tabular}
\end{table}

\begin{figure}[t]
\centering
\begin{tabular}{c}
    \includegraphics[width=80mm]{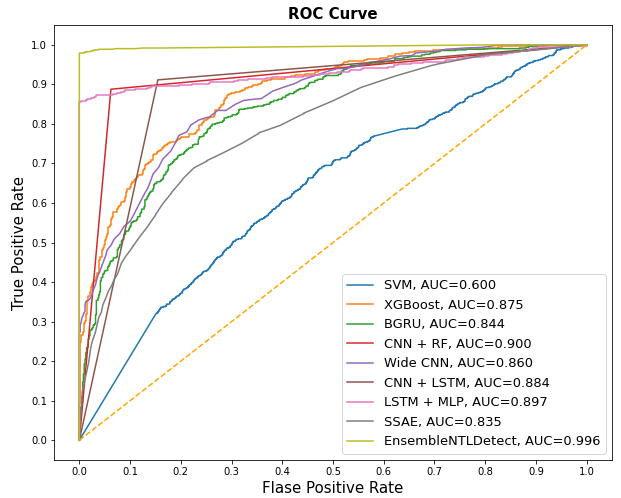}\\
\end{tabular}
\caption{ROC Curve of EnsembleNTLDetect Vs. state-of-the-art electricity theft detection models}
\label{auc_roc}
\end{figure}

Further, the complete set of experiments were repeated to validate the performance of EnsembleNTLDetect for different sizes of the train and test dataset (Table \ref{split_table}). The train-test split size of 50:50, 60:40, and 70:30 was used to demonstrate the impact of CTGAN on the overall performance of EnsembleNTLDetect. In all the cases, EnsembleNTLDetect 
provides a quality metric score of above 0.97, especially when the training data size is reduced to 60\% and 70\%, the application of CTGAN for synthetic sample generation nullifies the imbalanced nature of the dataset and provides AUC-ROC and MCC values of above 0.97.

Table \ref{cmp_table} provides a detailed comparative analysis of EnsembleNTLDetect over the state-of-the-art electricity theft detection models in terms of the considered quality metrics. Figure \ref{auc_roc} presents the AUC-ROC analysis of EnsembleDetect and the state-of-the-art electricity theft detection models. Even though SVMs are widely explored and applied in various forms for NTL detection, they provide marginal performance due to overfitting and high susceptibility to noise. Despite its benefits, such as memorization and generalization from the deep CNN architecture and wide components, Wide and deep CNN is ranked as an average model due to its inability to model long time series. CNN-LSTM architecture with CNNs as feature extractors and LSTMs to model long sequences in the time series data proves to be the best in the state-of-the-art electricity theft detection models. However, the complex architectures, high training time, and overfitting issues create a major impact while deploying in a real-time environment. In such cases, EnsembleNTLDetect provides optimal performance with minimal training time. Moreover, state-of-the-art NTL detection methods in \cite{b23, b48, b50} have not been verified on SGCC dataset, which partly explains their subpar performance on this dataset and lack of generalization ability. Table \ref{efficiency_table} shows the execution time (in Mins) for each component of EnsembleNTLDetect. It takes about an hour to deploy and execute the entire framework from scratch, while the tradeoff between efficiency and accuracy is perfectly managed by the time taken to generating predictions within few milliseconds. The overall computation time can be further reduced with high-end computing infrastructures at the smart grid control stations. The extensive experiments carried out using the SGCC dataset have resulted in the following observations,
\begin{enumerate}
     \item [1] eDTWBI works exceptionally well in all scenarios, except when the nearby consumption values are very low, resulting in 0 as consumption values. Further, the introduction of $Search\_Size$ parameter reduces the overall execution time of the DTW algorithm through limiting the search space of the search window.
    \item [2] Near-miss undersampling does not result in any loss of information which is indicated by a high recall score.
    \item [3] The training pipeline of stacked autoencoders was highly efficient and effective such that there was no loss of critical information even after reducing the dimensions by approximately 87\%. It also enables faster training and inference of the ML classifiers. 
    \item [4] Fine-tuning CTGAN aids the classifier to model all possible types of original and synthetic data, thereby enhancing the robustness of the model to completely unseen or aberrant data.
\end{enumerate}

\begin{table}[t]
\caption{Execution time analysis of EnsembleNTLDetect}
\label{efficiency_table}
    \centering
    \begin{threeparttable}
    \begin{tabular}{|l|l|}
    \hline
    \textbf{Components} & \makecell{\textbf{Execution time} \\ (in Mins)}\\
    \hline
    eDTWBI based imputation & 28.6\textsuperscript{*}\\
    \hline
    Z-Score for outlier removal & 1.2\\
    \hline
    Near miss undersampling (Version 1) & 6.7\\
    \hline
    \makecell[l]{Stacked Autoencoder:\\ (Training + generating latent vectors)} & 5.4\\
    \hline
    \makecell[l]{CTGAN:\\ (Training + generating 10,000 Samples)} & 13.4\\
    \hline
    \makecell[l]{Soft Voting Ensemble:\\ (Hyperparameter Tuning + training + prediction)} & 16.8\\
    \hline
    \textbf{Total execution time} & 72.1\\
    \hline
    \end{tabular}
    \begin{tablenotes}
        \item * Imputation was done locally
    \end{tablenotes}
    \end{threeparttable}
\end{table}

\section{Conclusions \& Future Work}\label{conclusion}

This paper presents EnsembleNTLDetect, a robust and scalable framework to detect electricity theft by analyzing the electricity consumption patterns of consumers. Specific contributions attributed to address the limitations in the state-of-the-art electricity theft detection models are, (i) \textbf{Consecutive missing values:} enhanced version of dynamic time warping algorithm imputes the large gaps in the time series data and seasonality trends were preserved by introducing $Search\_Size$ parameter to restrict the search space of the reference points within the season range of the missing values, (ii) \textbf{Imbalanced dataset:} near-miss undersampling technique generates the balanced dataset without information loss, (iii)\textbf{ High dimensional data:} stacked autoencoders with three autoencoders performs an unsupervised learning based dimensionality reduction on the feature space, (iv) \textbf{Efficient training:} a fine-tuned conditional GAN provides effective training for the classifiers through exposing them to real and synthetic data with different energy consumption patterns, and (v) \textbf{Effective classification:} a soft voting ensemble classification model that uses random forest and XGBoost learns the complex high dimensional electricity consumption patterns to detect the consumers with aberrant consumption patterns with high detection rate and less false alarm rate. Extensive experimental analysis on the SGCC real-time electricity consumption dataset demonstrates that EnsembleNTLDetect outperforms the state-of-the-art electricity theft detection models by accurately classifying genuine consumers and electricity thieves with a recall and MCC score of 0.98. Further, the application of stacked autoencoders based dimensionality reduction technique has reduced the total computational cost of EnsembleNTLDetect such that it ensures simple and effective deployment for large scale real-time electricity theft detection. Experiments and in-depth analysis of EnsembleNTLDetect on different open-source electricity consumption datasets for detecting NTLs in smart grids are planned as a future directive of this work. In addition, detailed analysis on the effects of consumer metadata on the consumption patterns requires more attention to understand the various social-psychological factors that impact the consumers' electricity consumption patterns.

\section{Acknowledgments}
We thank the Robert Bosch Centre for Data Science \& Artificial Intelligence, IIT-Madras; IIT-Bombay (Institute Postdoctoral Fellowship-AO/Admin-1/Rect/33/2019).

\end{document}